\pdfoutput=1

\documentclass[11pt]{article}
\usepackage{amssymb}

\usepackage[final]{acl}
\usepackage{booktabs}
\usepackage{multirow}
\usepackage{tcolorbox} 
\usepackage{times}
\usepackage{latexsym}
\usepackage[T1]{fontenc}

\usepackage[utf8]{inputenc}

\usepackage{microtype}

\usepackage{inconsolata}

\usepackage{graphicx}
\usepackage{array}
\usepackage{amsmath}
\usepackage{xcolor}
\title{The Missing Parts: Augmenting Fact Verification with Half Truth Detection}

\author{Yixuan Tang \qquad Jincheng Wang \qquad  Anthony K.H. Tung\\ \\
  School of Computing, National University of Singapore \\
  \texttt{yixuan@comp.nus.edu.sg, bertrand.wongjc@gmail.com,  atung@comp.nus.edu.sg} \\}

\usepackage{pifont}
\begin{document}
\maketitle
\begin{abstract}


Fact verification systems typically assess whether a claim is supported by retrieved evidence, assuming that truthfulness depends solely on what is stated. However, many real-world claims are \textit{half-truths}, factually correct yet misleading due to the omission of critical context. Existing models struggle with such cases, as they are not designed to reason about omitted information. We introduce the task of \textbf{half-truth detection}, and propose \textsc{PolitiFact-Hidden}, a new benchmark with 15k political claims annotated with sentence-level evidence alignment and inferred claim intent. To address this challenge, we present \textbf{TRACER}, a modular re-assessment framework that identifies omission-based misinformation by aligning evidence, inferring implied intent, and estimating the causal impact of hidden content. TRACER can be integrated into existing fact-checking pipelines and consistently improves performance across multiple strong baselines. Notably, it boosts Half-True classification F1 by up to 16 points, highlighting the importance of modeling omissions for trustworthy fact verification. The benchmark and code are available via \url{https://github.com/tangyixuan/TRACER}. 
\end{abstract}

\section{Introduction}

\begin{figure*}[t]
\centering
\includegraphics[width=0.88\linewidth]{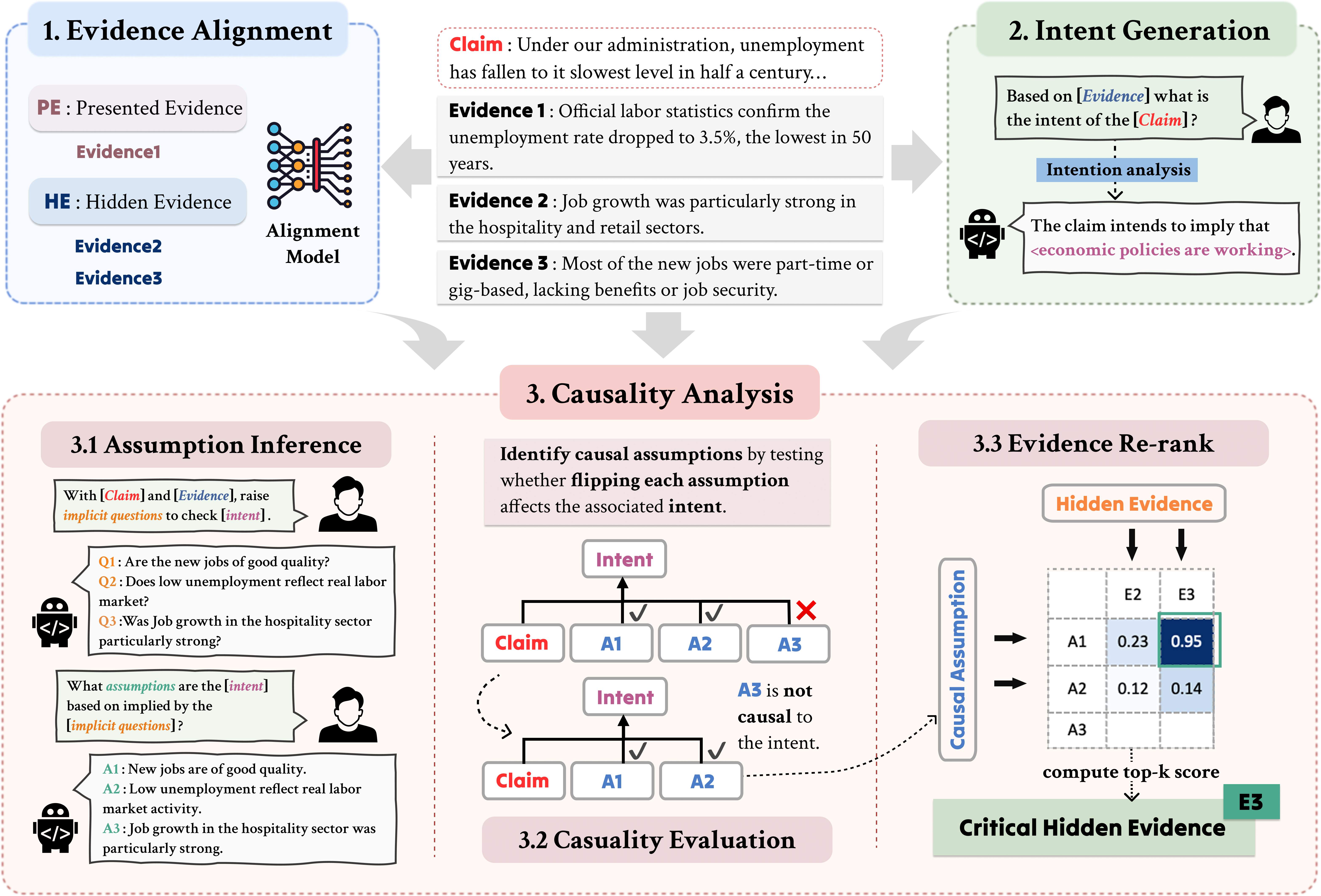}
\caption{Overview of the TRACER framework for half-truth detection. The system identifies Critical Hidden Evidence (CHE) through evidence alignment, intent generation, and causality analysis, and re-assesses claims for omission-based misinformation.}
\label{fig:framework}
\end{figure*}
 
\begin{table}[t]
\centering
\small
\begin{tabular}{@{}p{0.98\linewidth}@{}}
\toprule
\textbf{Claim:} Under our administration, unemployment has fallen to its lowest level in half a century, demonstrating that our economic policies are working. \\
\midrule
\textbf{Presented Evidence (PE):} \\
• Official labor statistics confirm the unemployment rate dropped to 3.5\%, the lowest in 50 years. \\
\midrule
\textbf{Hidden Evidence (HE):} \\
• Most of the new jobs were part-time or gig-based, lacking benefits or job security. $\rightarrow$ CHE \\
• Labor force participation remained low, with many discouraged workers no longer counted. $\rightarrow$ CHE \\
• Job growth was particularly strong in the hospitality and retail sectors.\\
\midrule
\textbf{Verdict by Standard FV Model:} \\
\textbf{True}: The claim is factually supported by official statistics. \\
\midrule
\textbf{TRACER Re-Assessment Verdict:} \\
\textbf{Half-True}: Although the unemployment figure is accurate, the omission of job quality and participation context distorts the implied economic success. \\
\bottomrule
\end{tabular}
\caption{A factually correct political claim re-evaluated as misleading (Half-True) by TRACER through Critical Hidden Evidence (CHE) analysis.}
\label{tab:eg}
\end{table}

The rapid spread of digital content has made fact verification a critical component in combating misinformation and promoting trustworthy public discourse. Traditional fact-checking systems follow a standard paradigm: given a claim and a body of evidence, the system classifies the claim as \textit{true}, \textit{false}, or \textit{not enough information}~\cite{survey_combating_misinformation}. These systems are effective in identifying clearly incorrect claims and continue to serve as the backbone of automated verification pipelines.

However, many real-world claims are not outright false but are still misleading due to the omission of critical context. Misinformation can evolve dynamically when propagated under different political stances ~\cite{DBLP:conf/emnlp/MPCG}, these are often referred to as \textbf{half-truths}, i.e. statements that are factually correct but strategically incomplete~\cite{beware_half_truth, cherry}. Consider the example in Table~\ref{tab:eg}, where a politician claims that unemployment has reached a 50-year low. While this statistic is factually accurate, it omits key information, such as the rise in part-time gig jobs and stagnant labor force participation, that undermines the implied narrative of broad economic success. Standard fact verification (FV) models, which focus on validating surface-level factuality, label such claims as \textit{true}, failing to capture the misleading nature of selective omission.

This challenge highlights a fundamental limitation in existing FV pipelines: they are not designed to reason about what is missing. Current models typically assess what is stated, treating veracity as a discrete property grounded in textual entailment~\cite{completeness, half_truth_deception}. Yet in practice, truthfulness is often shaped by both what is said and what is left unsaid. Omission-based misinformation exploits this gap, occupying a gray area between truth and falsehood that standard systems are ill-equipped to address.

In this paper, we introduce the task of \textbf{half-truth detection}, which complements traditional fact verification by modeling \textit{completeness}. We define half-truths as claims that are factually accurate but omit \textbf{Critical Hidden Evidence (CHE)}—information that, if included, would significantly alter the plausibility of the claim’s implied meaning. Our goal is to identify such omissions and assess their impact on the inferred intent of the claim.

To tackle this task, we propose \textbf{TRACER} (\textit{Truth ReAssessment with Critical Hidden Evidence reasoning}), a framework to augment fact-checking systems with omission-aware reasoning. TRACER operates in three stages: (1) \textbf{evidence alignment}, to classify retrieved evidence as presented or hidden; (2) \textbf{intent generation}, to recover the claim's implicit message; and (3) \textbf{causality analysis}, to determine whether the Hidden Evidence undermines the inferred intent. These components feed into a lightweight \textbf{re-assessment module} that revisits claims, particularly those initially labeled as \textit{true}, and identifies misleading omissions. TRACER is model-agnostic and can be integrated into both agent-based and prompting-based FV pipelines.

To support this task, we construct \textbf{\textsc{PolitiFact-Hidden}}, a benchmark dataset based on the PolitiFact corpus. It contains about 15k claims annotated with sentence-level labels indicating Presented and Hidden Evidence, along with inferred claim intents validated through a combination of LLM prompting and human quality control. To our knowledge, this is the first dataset to explicitly annotate both omission and intent, enabling systematic study of half-truths at scale.

\noindent Our contributions are as follows:
\begin{enumerate}
    \item We \textbf{formulate half-truth detection} as a new task in fact verification, targeting claims that omit critical context while remaining factually correct.
    \item We introduce \textbf{\textsc{PolitiFact-Hidden}}, a large-scale benchmark with fine-grained annotations for Presented / Hidden Evidence and inferred claim intent.
    \item We propose \textbf{TRACER}, a three-stage framework that identifies omission-based misinformation through evidence alignment, intent modeling, and causal reasoning. TRACER can be deployed as a re-assessment module and yields substantial gains in detecting half-truths across multiple strong baselines.
\end{enumerate}

By modeling completeness alongside correctness, this work advances the frontier of fact verification. It addresses a blind spot in current systems and offers a generalizable framework for uncovering more subtle forms of misinformation that operate through omission rather than distortion.

\section{Related Work}

\paragraph{Fact Verification.}
Fact verification is commonly framed as a three-stage pipeline involving claim detection, evidence retrieval, and claim classification into \textit{Supported}, \textit{Refuted}, or \textit{Not Enough Information}~\cite{thorne-etal-2018-fever,pipeline-survey}. Benchmarks such as FEVER~\cite{thorne-etal-2018-fever} and LIAR~\cite{wang-2017-liar} have facilitated significant progress in this area. Most existing systems focus on surface-level factual correctness, aiming to match claims against retrieved facts. While effective for outright falsehoods, these approaches are less suited to handling omission-driven manipulation.

\paragraph{Omission and Half-Truths.}
Omission-based misinformation, including half-truths, has received increasing attention. \citet{beware_half_truth} introduce controlled claim editing to expose omitted content, and \citet{implicit-question} propose generating implicit questions to recover missing context. Other datasets have incorporated related annotations, such as \textit{Cherry-picking}~\cite{schlichtkrull2023averitec} and \textit{Mixture}~\cite{rawfc-yang-etal-2022-coarse}, which primarily capture conflicting evidence rather than omissions per se. These schemes focus on factual inconsistency (i.e., presence of both supporting and refuting evidence), rather than semantic incompleteness or intent-driven distortion. In contrast, our work targets \textit{half-truths}, claims that are factually accurate but strategically omit Critical Hidden Evidence (CHE) that significantly alters interpretation. Closely related are efforts that explore the role of intent in misinformation, such as distinguishing disinformation through concealment and overstatement~\cite{definition,overstatement_concealment}. \citeauthor{DBLP:conf/emnlp/NEWSCOPE} (\citeyear{DBLP:conf/emnlp/NEWSCOPE}) uncover the comprehensive view of events by mitigating selective presentation of information, they do not integrate downstream fact verification. We go beyond these by explicitly modeling the causal impact of Hidden Evidence on inferred intent without altering the original claim.

\paragraph{Reasoning-Based Fact Checking.}
Recent methods incorporate structured reasoning to improve factuality assessment. Program-guided models such as QACheck and ProgramFC~\cite{pan-etal-2023-fact} generate intermediate steps to support verification~\cite{DBLP:conf/eacl/TangNT21}. Argumentation-based approaches, such as CHECKWHY~\cite{si-etal-2024-checkwhy}, model causal links within evidence chains. Meanwhile, prompting-based methods like HiSS~\cite{hiss} and Flan-T5~\cite{flan-t5} leverage large language models for step-by-step verification. Other work explores intent modeling using contrastive learning~\cite{intent-matrix} or refined retrieval~\cite{intent-query}. Our work complements these efforts by introducing omission-aware reasoning and providing a modular framework that can be integrated into both structured and generative pipelines.

\section{Task Formulation}

We define \textbf{half-truth detection} as an extension of fact verification that focuses on \textit{factual completeness}. A claim may be factually accurate in isolation, yet convey a misleading impression by omitting relevant information that influences its interpretation. The goal is to identify such omissions and assess whether they materially affect the plausibility of the claim’s implied message.

Formally, given a claim $C$ and a set of retrieved evidence sentences $E = \{e_1, e_2, \dots, e_n\}$ relevant to $C$, the goal is to classify the claim into one of three categories: \textit{True}, \textit{Half-True}, or \textit{False}. This classification is determined not only by factual support but also by the presence or absence of \textbf{Critical Hidden Evidence (CHE)} $\subseteq E$ that is both (1) not presented in the claim, and (2) necessary to understand or challenge the claim’s implied conclusion.

To support this, we define the following components:

\begin{itemize}
    \item \textbf{Presented Evidence (PE)}: Sentences in $E$ that are explicitly stated or clearly implied in the claim.
    \item \textbf{Hidden Evidence (HE)}: Sentences in $E$ that are relevant to the claim but not mentioned.
    \item \textbf{Intent}: The implied conclusion or message that the claim is likely to convey to the reader.
    \item \textbf{Critical Hidden Evidence (CHE)}: A subset of HE that, if revealed, would significantly affect the plausibility of the claim’s intent.
\end{itemize}

\begin{table}[t]
\small
  \centering
  \begin{tabular}{l | l}
    \toprule
    \textbf{Consolidated Label} & \textbf{Original Rating(s)} \\
    \midrule
    True & True \\
    Half-True & Mostly True, Half-True \\
    False & Mostly False, False, Pants on Fire \\
    \bottomrule
  \end{tabular}
  \caption{Mapping from original PolitiFact ratings to consolidated labels.}
  \label{tab:politifact_rating}
\end{table}

\begin{table}[t]
\small
\centering
\begin{tabular}{c|ccc|c}
\toprule
\textbf{Split} & \textbf{True} & \textbf{Half-True} & \textbf{False} & \textbf{Total} \\
\midrule
\textbf{Train} & 1,352 & 4,564 & 6,078 & 11,994 \\
\textbf{Dev}   & 64    & 195   & 741   & 1,000  \\
\textbf{Test}  & 93    & 406   & 1,501 & 2,000  \\
\bottomrule
\end{tabular}
\caption{Distribution of labels in the \textsc{PolitiFact-Hidden} dataset across train/dev/test splits.}
\label{tab:data_distribution}
\end{table}

This formulation connects closely to the traditional FV pipeline but adds a new layer of reasoning: not only must a system verify what is said, it must also reason about what is left unsaid. By focusing on omissions that shift the meaning of a claim, half-truth detection supports a more nuanced understanding of misinformation and helps uncover subtle forms of manipulation that standard FV systems may overlook.


\section{Dataset: \textsc{PolitiFact-Hidden}}
\label{sec:dataset}

\begin{figure}[t]
  \includegraphics[width=0.9\columnwidth]{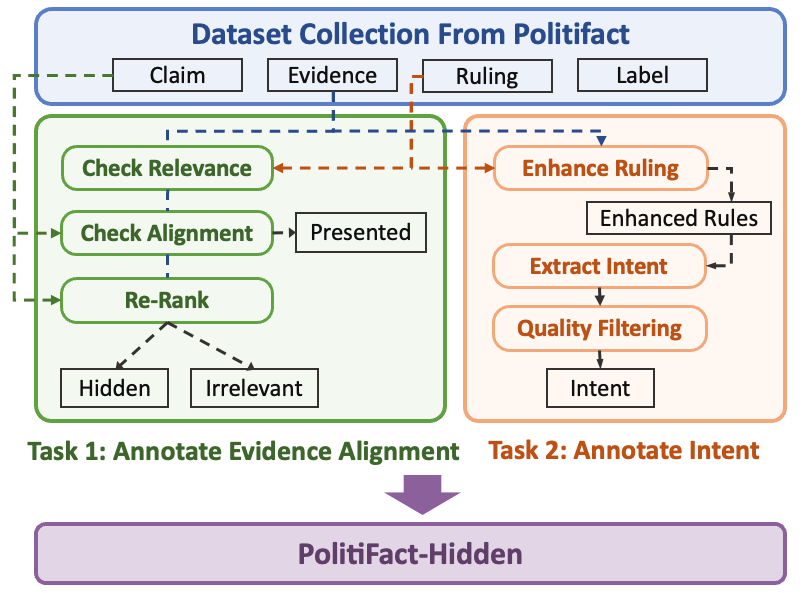}
  \caption{Illustration of the semi-automated annotation pipeline for constructing PolitiFact-HIDDEN, combining \texttt{GPT-4o-mini} prompting with human quality control.}
  \label{fig:annotation}
\end{figure}

As illustrated in Figure~\ref{fig:annotation}, we develop a semi-automated annotation pipeline (Figure~\ref{fig:annotation}) combining \texttt{GPT-4o-mini} prompting and model-assisted refinement to label each claim with evidence alignment and Intent.

\begin{table*}[t]
  \centering
  \small
  \begin{tabular}{l|l|c|c|c}
    \toprule
    \textbf{Dimension} & \textbf{Requirement} & \textbf{LLM-Positive} & \textbf{Human Confirmed} & \textbf{Agreement} \\ \midrule
    \textbf{Plausibility} &  The inferred intent must not contradict the claim. & 95 & 94 & 98.9\% \\
    \textbf{Implicity}    & The intent should be implied, not overtly stated. & 94 & 93 & 98.9\% \\
    \textbf{Sufficiency}  & The description must be specific and informative. & 81 & 80 & 98.8\% \\
    \textbf{Readability}  & The intent must be clearly and fluently expressed. & 76 & 70 & 92.1\% \\
    \bottomrule
  \end{tabular}
  \caption{Agreement between LLM and human annotations across intent quality dimensions.}
  \label{tab:agreement_selection}
\end{table*}

We introduce \textsc{PolitiFact-Hidden}, a benchmark for omission-aware fact verification. It extends the original PolitiFact corpus with fine-grained annotations capturing both Presented and Hidden Evidence, and the Intent behind each claim. These annotations enable systematic evaluation of whether omitted content, i.e. Critical Hidden Evidence (CHE), alters the claim’s implied meaning.

\subsection{Data Source and Label Schema}

The dataset is built upon fact-checking articles from PolitiFact, which include both a concise claim and an accompanying verdict article. Unlike many other fact-checking sources, PolitiFact explicitly considers completeness in its rating criteria: a claim rated \textit{True} must be both accurate and complete, while \textit{Mostly True} and \textit{Half-True} indicate factual correctness with missing context \citep{politifact_principles}. In contrast, \textit{Mostly False} reflects the presence of conflicting evidences.

We consolidate PolitiFact’s original six-level rating into three coarse-grained labels to align with our half-truth detection task:

Each article is split into evidence paragraphs, which provide factual context, and ruling paragraphs, which justify the final verdict. To prevent label leakage, we separate these segments using structural cues (e.g., ``Our Ruling'') and exclude ruling content from model input.

To improve generalization and test temporal robustness, we collect an additional 2,000 claims from 2020–2025 to form a temporally disjoint test set. Claims with date overlap are removed from the training pool. The resulting dataset contains 14,994 claims. Detailed statistics are shown in Table~\ref{tab:data_distribution}.

\subsection{Annotation Pipeline}

\label{sec:anno}

\paragraph{Evidence Annotation}  
For each evidence sentence, we determine whether it is already reflected in the claim. This involves:
\begin{enumerate}
    \item  \textbf{Relevance Check:} Filter out irrelevant content using LLM-based entailment prompting.
    \item  \textbf{Presentation Check:} Assess whether the content is explicitly or implicitly stated in the claim.
    \item  \textbf{Similarity Refinement:} Use cosine similarity with XLM-RoBERTa embeddings\cite{similarity_ranking_model} to refine edge cases and mitigate hallucinations.
\end{enumerate}

Evidence is labeled as either PE or HE. Manual inspection of 50 samples showed an 88\% agreement between LLM predictions and human judgments, validating the alignment process.

\paragraph{Intent Annotation.}
A key element of half-truth detection is the claim’s Intent, i.e., the implied message or judgment it seeks to convey. Intents are extracted in 3 steps:
\begin{enumerate}
    \item  \textbf{Ruling Enhancement:} Enhance ruling text by adding supporting evidence for clarity.
    \item   \textbf{Intent Extraction:} Use instruction-tuned prompting to extract the claim’s intended conclusion.
    \item  \textbf{Quality Filtering:} Filter extracted intents using four criteria, namely plausibility, implicity, sufficiency and readability.
\end{enumerate}
To validate the quality of LLM-based filtering, we had two human annotators independently assess 100 samples across the same four evaluation dimensions. Agreement between the LLM and both annotators was high (92.1-98.9\% across dimensions), suggesting that the LLM-assisted approach reliably captures high-quality intents for downstream reasoning. The full intent evaluation prompts are provided in Appendix~\ref{appendix:data}.

\section{The TRACER Framework}
\label{sec:method}

We propose \textbf{TRACER}, a modular framework for detecting half-truths by identifying and evaluating omitted context. TRACER is designed to integrate with existing fact verification (FV) systems by reassessing claims, particularly those initially labeled as \textit{True}, to determine whether omissions materially alter the claim’s intended message.

TRACER operates in three stages: (1) evidence alignment, (2) intent generation, and (3) causal estimation of omitted content. These components support a final re-assessment module that refines the output of base FV models.

\subsection{Evidence Alignment}
\label{sec:alignment}

The first stage determines whether each evidence sentence $e_i \in E$ is explicitly or implicitly reflected in the claim $C$. We formulate this as a binary classification task, assigning each $e_i$ to either Presented Evidence (PE) or Hidden Evidence (HE). Only HE is forwarded for further analysis.

\begin{figure}[t]
  \centering
  \includegraphics[width=0.8\columnwidth]{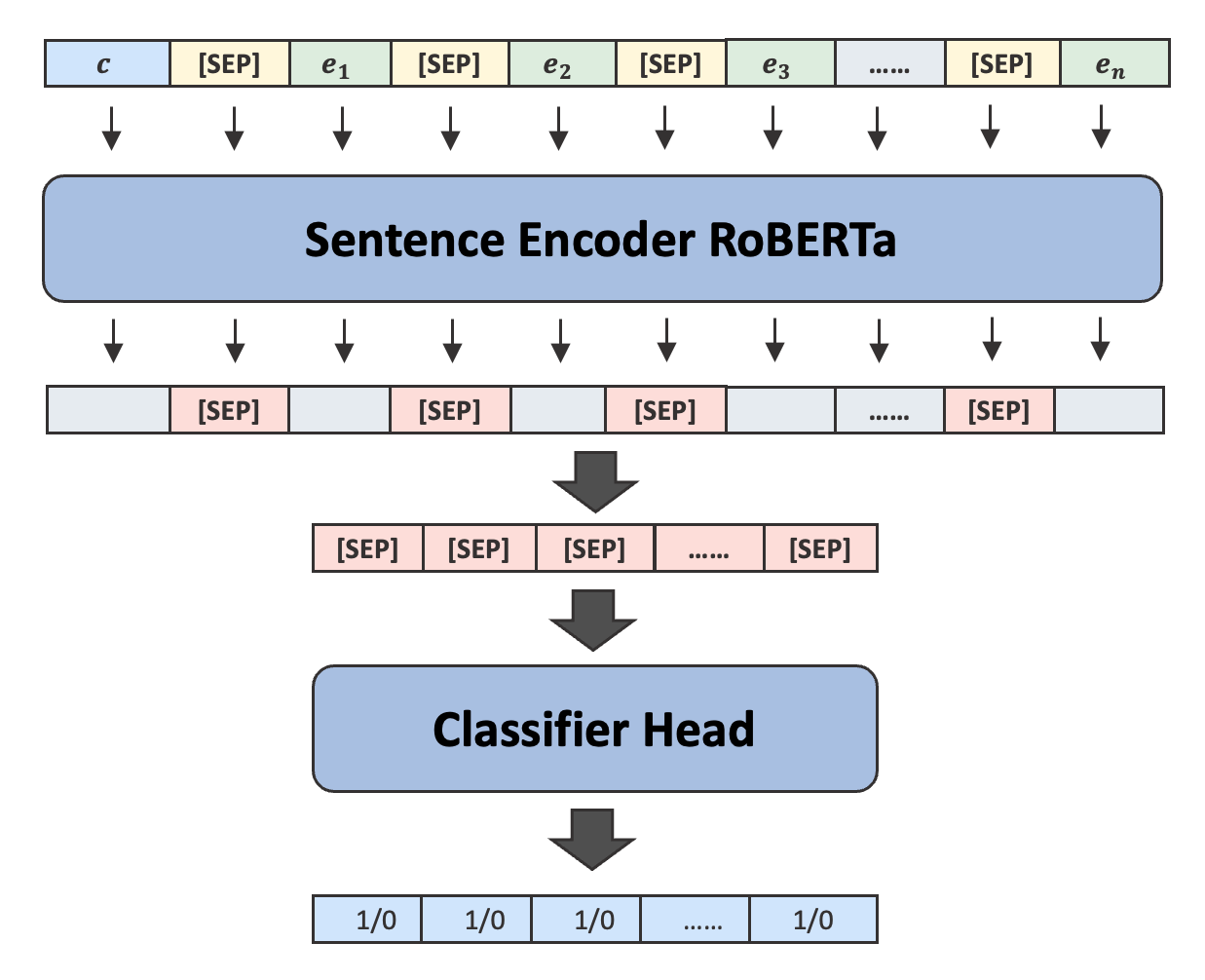}
  \caption{Architecture of the evidence alignment module, which classifies each evidence sentence as presented or hidden relative to the claim.}
    \label{fig:alignment_model}
\end{figure}

As shown in Figure~\ref{fig:alignment_model}, a transformer-based alignment model is adopted. Each $(C, e_i)$ pair is concatenated and encoded using RoBERTa-large~\citep{roberta-large}. A classification head predicts whether the evidence content is present in the claim. This alignment step enables TRACER to isolate potentially omitted but relevant information for downstream intent and causal reasoning.

\subsection{Intent Generation}
\label{sec:intent}

Understanding this latent intent is essential for determining whether omitted content is misleading. As described in Section \ref{sec:anno}, we prompt-tune an LLM using input that includes the claim and its associated evidence context to infer intent. This prompt-based formulation encourages the model to extract implicit conclusions without relying on manually predefined templates. The resulting intents serve as semantic anchors for subsequent causality analysis.

\subsection{Causality Analysis}
\label{sec:causality}

While assumptions are derived from HE, not all HE sentences directly affect the plausibility of the intent. Many are tangential or neutral. To distinguish Critical Hidden Evidence (CHE) from neutral omissions, we estimate the causal influence of each HE sentence on the inferred intent.

Inspired by abductive reasoning frameworks~\cite{implicit-question}, we generate candidate assumptions $A_i$ that must hold for the intent $Z$ to be valid. These assumptions are derived from evidence through binary question generation and abstraction.

We then evaluate the impact of each $A_i$ using counterfactual prompting: given $do(A_i = \neg A_i)$, does the intent $Z$ still hold? If not, $A_i$ is marked as causally important. For each validated assumption, we retrieve corresponding CHE from the HE pool by selecting sentences that either support or contradict it, based on semantic similarity and an NLI model that verifies logical entailment. This two-step refinement prevents irrelevant or weakly related evidence from being misclassified as CHE.

\subsection{Final Re-Assessment Module}
\label{sec:reassessment}

To determine the final label (\textit{True}, \textit{Half-True}, or \textit{False}), we incorporate the inferred intent, assumptions, and selected CHE into a re-assessment module (RA). This module re-evaluates the original FV prediction, especially when the claim was initially classified as \textit{True}.

If no CHE is found, the original label is preserved. If CHE alters the plausibility of the intent, the system reclassifies the claim as \textit{Half-True} or \textit{False}, depending on the nature of the conflict. This re-assessment stage is implemented as a prompt-based module. It is designed to be model-agnostic and can be plugged into existing FV pipelines to enhance their ability to detect omission-based manipulation. We provide the full prompt examples used in each component of TRACER in Appendix~\ref{sec:appendix_prompts}.

\begin{figure}[t]
    \centering
    \includegraphics[width=\columnwidth]{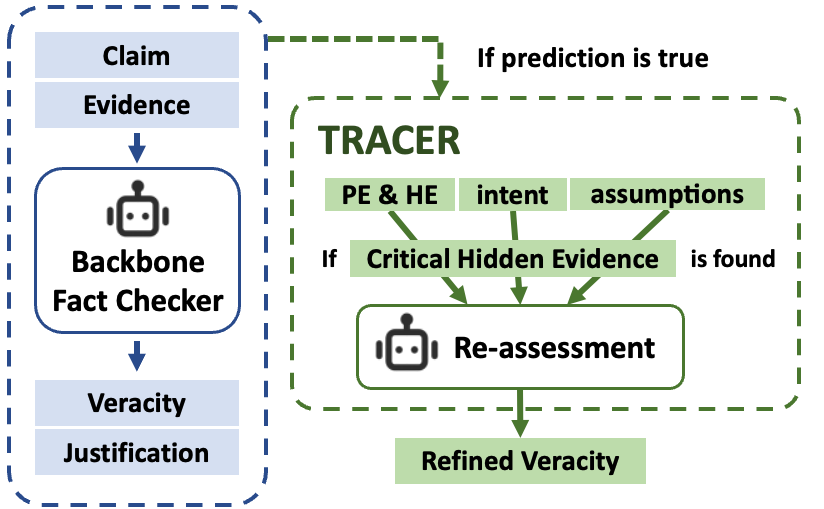}
     \caption{TRACER integrated into a fact verification pipeline as a re-assessment module.}    
    \label{fig:reassessment}
\end{figure}

\section{Experiments}

We evaluate TRACER by integrating it into existing fact verification (FV) models and measuring its effectiveness in identifying omission-based misinformation. Specifically, we compare TRACER-enhanced models against strong FV systems and conduct ablation studies to assess the impact of individual components. Evaluation metrics include overall Accuracy, macro-F1, and F1 on the \textit{Half-True} class (F1(H)), which reflects the system's ability to capture omission-driven misinterpretations.

\subsection{Evidence Alignment}

We train our evidence alignment model using \texttt{RoBERTa-large}\footnote{\url{https://huggingface.co/FacebookAI/roberta-large}}, with a context-aware batch sampling strategy. At each training step, sequential evidence segments are grouped into a batch to help the model leverage intra-batch contextual signals. We compare this setup to a baseline where each claim-evidence pair is processed independently (i.e., context-unaware). Both models are trained for 5 epochs with a batch size of 8 and a learning rate of 1e-5.

\begin{table}[th]
    \centering
    \small
    \begin{tabular}{l | cc}
        \toprule
        \textbf{Method} & \textbf{Accuracy} & \textbf{F1} \\
        \midrule
        \textbf{RoBERTa-large}  & 93.2 & 90.3 \\
        \textbf{TRACER} (context-aware) & \textbf{94.0} & \textbf{91.6} \\
        \bottomrule
    \end{tabular}
    \caption{Evidence alignment performance.}
    \label{tab:evidence_alignment_res}
\end{table}

As shown in Table~\ref{tab:evidence_alignment_res}, the context-aware training improves F1 by 1.3 and accuracy by 0.8, showing enhanced ability to detect omitted evidence.

\subsection{Intent Generation}

We fine-tune \texttt{GPT-4o-mini} via the OpenAI API to generate implicit intent statements. Each training input includes the claim and relevant evidence paragraphs. We compare this approach to a 4-shot in-context prompting baseline. The fine-tuned model is trained for 3 epochs with a batch size of 4.

\begin{table}[th]
    \centering
    \small
    \begin{tabular}{l | ccc}
        \toprule
        \textbf{Method} & \textbf{ROUGE-L} & \textbf{BLEU} & \textbf{BERTScore} \\
        \midrule
        \textbf{Few-shot} & 37.7 & 6.1 & 91.2 \\
        \textbf{Fine-tuned} & \textbf{46.2} & \textbf{8.0} & \textbf{91.5} \\
        \bottomrule
    \end{tabular}
    \caption{Performance of intent generation methods.}
    \label{tab:intent_generation_res}
\end{table}

As shown in Table~\ref{tab:intent_generation_res}, fine-tuning consistently outperforms prompting across all metrics, supporting our decision to use supervised intent extraction in TRACER.

\begin{table}[t]
\centering
\small
\begin{tabular}{l | cc | cc} 
\toprule
\multirow{2}{*}{\textbf{Method}} & \multicolumn{2}{c|}{\textbf{Accuracy}} & \multicolumn{2}{c}{\textbf{F1}} \\ \cmidrule{2-3}\cmidrule{4-5}
               & Dev & Test & Dev & Test \\ \midrule
\textbf{QACheck}         & 48.5  & 48.8  & 38.0  & 38.6  \\
\textbf{ProgramFC}       & 55.4  & 56.9  & 32.9  & 34.2  \\
\textbf{CHECKWHY}        & 74.8  & 65.9  & 64.2  & 54.6  \\
\textbf{Flan-T5}         & 69.7  & 70.0  & 50.8  & 50.4  \\ \midrule
\textbf{CoT}             & 76.6  & 76.3  & 68.5  & 64.3  \\
\textbf{CoT} \textsc{+RA}  & 77.3  & 78.5  & \textbf{68.7}  & \textbf{68.0}  \\
\quad \textit{Improvement} & 
\textit{\small \textcolor{blue}{$\uparrow$ 0.7}} & 
\textit{\small \textcolor{blue}{$\uparrow$ 2.2}} & 
\textit{\small \textcolor{blue}{$\uparrow$ 0.2}} & 
\textit{\small \textcolor{blue}{$\uparrow$ 3.7}} \\

\midrule
\textbf{HiSS}             & 76.3  & 78.3  & 60.3  & 59.4  \\
\textbf{HiSS} \textsc{+RA} & \textbf{78.1}  & \textbf{81.9}  & 64.3  & 65.7  \\
\quad \textit{Improvement} & 
\textit{\small \textcolor{blue}{$\uparrow$ 1.8}} & 
\textit{\small \textcolor{blue}{$\uparrow$ 3.6}} & 
\textit{\small \textcolor{blue}{$\uparrow$ 4.1}} & 
\textit{\small \textcolor{blue}{$\uparrow$ 6.3}} \\
\bottomrule
\end{tabular}
\caption{Overall accuracy and macro-F1 on fact verification. RA denotes integration of the TRACER re-assessment module.}
\label{tab:overall_results}
\end{table}

\begin{table*}[t]
\small
\centering
\begin{tabular}{l | ccc | ccc}
\toprule
\multirow{2}{*}{\textbf{Method}} & \multicolumn{3}{c|}{\textbf{Dev}} & \multicolumn{3}{c}{\textbf{Test}} \\ \cmidrule(r){2-7}
               & \textbf{Precision} &\textbf{ Recall} & \textbf{F1} & \textbf{Precision} &\textbf{ Recall} & \textbf{F1} \\
\midrule
\textbf{QACheck}       & 24.0 & 55.1 & 33.4 & 24.3 & 54.4 & 33.6 \\
\textbf{ProgramFC}     & 11.8 & 2.0  & 3.5  & 18.2 & 6.9  & 10.0 \\
\textbf{CHECKWHY}      & 43.2 & 76.5 & 55.3 & 34.3 & 58.1 & 43.1 \\
\textbf{Flan-T5}       & 37.7 & 33.7 & 35.6 & 44.1 & 27.3 & 33.7 \\ \midrule
\textbf{CoT}             & 44.6 & 71.8 & 55.0 & 45.0 & 63.8 & 52.8 \\
\textbf{CoT} \textsc{+RA}  & 45.5 \textit{\small \textcolor{blue}{$\uparrow$ 0.9}} & \textbf{83.6} \textit{\small \textcolor{blue}{$\uparrow$ 11.8}} & \textbf{59.0} \textit{\small \textcolor{blue}{$\uparrow$ 4.0}} & 48.5 \textit{\small \textcolor{blue}{$\uparrow$ 3.5}} & \textbf{79.3}  \textit{\small \textcolor{blue}{$\uparrow$ 15.5}} & 60.2 \textit{\small \textcolor{blue}{$\uparrow$ 7.4}} \\  \midrule

\textbf{HiSS}             & 34.9 & 47.6 & 40.2 & 53.7 & 37.9 & 44.4 \\
\textbf{HiSS} \textsc{+RA} & \textbf{46.3} \textit{\small \textcolor{blue}{$\uparrow$ 11.4}} & 54.4 \textit{\small \textcolor{blue}{$\uparrow$ 6.8}}  & 50.0 \textit{\small \textcolor{blue}{$\uparrow$ 9.8}} & \textbf{55.3} \textit{\small \textcolor{blue}{$\uparrow$ 1.6}} & 66.8  \textit{\small \textcolor{blue}{$\uparrow$ 28.9}}  & \textbf{60.5} \textit{\small \textcolor{blue}{$\uparrow$ 16.1}}  \\
\bottomrule
\end{tabular}
\caption{Precision, Recall, and F1 on the Half-True category. TRACER consistently improves detection of omission-based manipulation across all backbones.}
\label{tab:half_true_results}
\end{table*}

\subsection{Baselines}

TRACER requires the fact-checking method to produce justifications for the claim’s veracity. This is because TRACER assesses truthfulness by jointly considering the factual accuracy of the claim and the plausibility of its intent, where the former should be supported by explicit reasoning steps. We evaluate TRACER on top of two leading fact verification models that are suitable for integration:
\begin{itemize}
    \item \textbf{Chain-of-Thought (CoT)}~\cite{zeroshot-cot}: a zero-shot prompting baseline, where the model is guided to generate intermediate reasoning steps before producing the final fact-checking verdict.
    \item \textbf{HiSS}~\cite{hiss}: a state-of-the-art instruction-following verifier that employs structured reasoning by decomposing the claim into multiple verifiable subclaims and evaluating them step by step.
\end{itemize}

We also report results for the following four baselines:
\begin{itemize}
    \item \textbf{QACheck}~\cite{qacheck} and \textbf{ProgramFC}~\cite{pan-etal-2023-fact}: agent-based fact-checkers. QACheck decomposes claims into sub-questions and verifies them with evidence. ProgramFC treats verification as a structured program of sub-tasks generated via in-context learning and executed by modular agents.
    \item \textbf{CHECKWHY}~\cite{si-etal-2024-checkwhy} and \textbf{Flan-T5}~\cite{flan-t5}: prompting-based LLMs. CHECKWHY models causal reasoning through argument structures. Flan-T5 is identified as a strong fact verifier in hallucination evaluations.

\end{itemize}

To ensure fairness, we evaluate all baselines using \texttt{GPT-4o-mini}, except in cases where prior work demonstrates that a different backbone yields stronger performance. For HiSS, we find \texttt{GPT-3.5-turbo} consistently outperforms \texttt{GPT-4o-mini}.

\subsection{Main Results}       
We present the overall performance of TRACER-integrated models and baselines in Table~\ref{tab:overall_results} (Accuracy and macro-F1) and Table~\ref{tab:half_true_results} (Precision, Recall, and F1 on the Half-True category). The results highlight TRACER’s consistent improvements in both general fact verification and the more challenging omission-sensitive cases.

\paragraph{Overall Performance.}
Table~\ref{tab:overall_results} shows that TRACER improves both accuracy and macro-F1 when added to strong reasoning-based backbones. For example, integrating TRACER with HiSS improves test accuracy from 78.3\% to 81.9\%, and macro-F1 from 59.4 to 65.7. Similarly, CoT benefits from TRACER with a 2.2 point gain in test accuracy and a 3.7-point increase in macro-F1. These gains are observed across both dev and test sets, indicating the robustness of TRACER as a general-purpose re-assessment module.

\paragraph{Half-True Detection.}
As shown in Table~\ref{tab:half_true_results}, TRACER substantially enhances performance on the Half-True class. When applied to HiSS, TRACER improves F1 by 16.1 points on the test set (from 44.4 to 60.5) and recall by 28.9 points (from 37.9 to 66.8), demonstrating its effectiveness in identifying omission-based manipulation. Similar improvements are seen for CoT, with F1 increasing from 52.8 to 60.2 and recall rising by 15.5 points (from 63.8 to 79.3).

Agent-based baselines such as QACheck and ProgramFC achieve low recall and F1, highlighting their inability to capture hidden context. In contrast, prompting-based methods are more competitive, but still benefit significantly from TRACER's re-assessment. These results validate our hypothesis that omission-aware reasoning, grounded in evidence alignment, intent modeling, and causal analysis, substantially improves a model’s ability to detect half-truths.

\paragraph{Per-Class Performance.}
To further examine TRACER's effect on fact verification, we report per-class performance for the top-performing models. As shown in Table~\ref{tab:per-class}, TRACER substantially improves the classification of \textit{Half-True} claims while also maintaining or slightly enhancing performance on \textit{True} and \textit{False} claims. This confirms that the observed gains are not achieved at the expense of other classes. 

\paragraph{Generalization.}

To examine the generalization of TRACER, we evaluate it with the open-source \texttt{LLaMA2-7B} model as the base verifier on the top-performing HiSS framework. With TRACER, accuracy improves from 78.2 to 82.3 and Macro-F1 from 59.1 to 65.4.
A breakdown of per-class performance and a follow-up analysis of results over different claim lengths is provided in Appendix~\ref{app:llama2}.

\begin{table}[t]
\centering
\small
\begin{tabular}{l | ccc} 
\toprule

 \textbf{Method} & \textbf{True} & \textbf{Half-True} & \textbf{False} \\ \midrule
\textbf{CoT}             & 52.9  & 52.8  & \textbf{87.1}  \\
\textbf{CoT} \textsc{+RA} & \textbf{56.7} \textit{\small \textcolor{blue}{$\uparrow$ 3.8}} & 
                            \textbf{60.2} \textit{\small \textcolor{blue}{$\uparrow$ 7.4}} & 
                            \textbf{87.1} \textit{\small (–)} \\

\midrule
\textbf{HiSS}             & 44.7  & 44.4  & 88.9  \\
\textbf{HiSS} \textsc{+RA} & \textbf{46.6} \textit{\small \textcolor{blue}{$\uparrow$ 1.9}} & 
                            \textbf{60.5} \textit{\small \textcolor{blue}{$\uparrow$ 16.1}} & 
                            \textbf{90.1} \textit{\small \textcolor{blue}{$\uparrow$ 1.2}} \\
\bottomrule
\end{tabular}
\caption{Per-class F1 scores on the test set.}
\label{tab:per-class}
\end{table}

\subsection{Qualitative Analysis}

To illustrate how TRACER detects omission-based manipulation, we present representative examples from the \textsc{PolitiFact-Hidden} test set. These cases show how factually accurate claims can still mislead through selective presentation, and how TRACER corrects such misclassifications by identifying Critical Hidden Evidence (CHE) and reasoning about intent.

\paragraph{Example: Misleading Attribution of Rising Costs.}
\textit{Claim: “Under the Obama economy, utility bills are higher.”}  
This claim was labeled \textbf{True} by HiSS, as it aligns with data showing an increase in utility costs during President Obama’s term. However, TRACER inferred an intent to attribute blame for rising prices to Obama’s economic policies. It then retrieved CHE showing that electricity prices rose even faster under the previous administration and followed a similar pattern across presidencies. This weakened the implied causal attribution and led TRACER to revise the label to Half-True.

Retrieved CHE:
\textit{“Rates rose at a significantly faster pace under Bush than they did under Obama.”}  
\textit{“Trends were not radically different between the Bush and Obama administrations.”}

\subsection{Ablation Study}

\begin{table}[t]
\centering
\small
\begin{tabular}{c | ccc | cc}
\toprule
\textbf{Cfg} & \textbf{Intent} & \textbf{Assump.} & \textbf{Causal.} & \textbf{F1 (H)} & \textbf{F1} \\
\midrule
\textcircled{\raisebox{-0.2ex}{1}} & -- & -- & -- & 44.4 & 59.4 \\
\textcircled{\raisebox{-0.2ex}{2}} & \checkmark & -- & -- & 50.9 & 64.7 \\
\textcircled{\raisebox{-0.2ex}{3}} & \checkmark & \checkmark & -- & \textbf{61.2} & 61.7 \\
\textcircled{\raisebox{-0.2ex}{4}} & \checkmark & \checkmark & \checkmark & 60.5 & \textbf{65.7} \\
\bottomrule
\end{tabular}
\caption{Ablation results for TRACER components.}
\label{tab:ablation_study}
\end{table}

We conduct an ablation study to evaluate the contribution of each component within the TRACER framework. Using HiSS as the base verifier, we progressively introduce intent modeling, assumption inference, and causality estimation. Results are shown in Table~\ref{tab:ablation_study}.

\paragraph{Impact of Intent Modeling.}

Setting \textcircled{\raisebox{-0.2ex}{1}} represents the base HiSS model without any TRACER components. In Setting \textcircled{\raisebox{-0.2ex}{2}}, we introduce intent generation but omit assumption inference and causality estimation. CHE is retrieved directly based on the inferred intent. This setup yields a substantial improvement in both F1(H) and macro-F1, rising from 44.4 to 50.9 and from 59.4 to 64.7, respectively, demonstrating that intent modeling alone provides meaningful signals for identifying omission-based misdirection.

\paragraph{Assumption Inference.}
Setting \textcircled{\raisebox{-0.2ex}{3}} extends the previous configuration by incorporating assumption inference, where the inferred intent is decomposed into finer-grained, testable assumptions. However, causality estimation is still disabled in this setting, meaning that all generated assumptions are treated equally during CHE retrieval. This leads to a further boost in F1(H) to 61.2, validating the utility of breaking down intent into more specific reasoning units. Nonetheless, macro-F1 decreases slightly to 61.7 due to an increase in false positives, indicating that not all assumptions contribute constructively.

\paragraph{Causality Filtering.}
In Setting \textcircled{\raisebox{-0.2ex}{4}}, our full TRACER framework is applied, with all components enabled, including causality estimation to filter out non-causal or spurious assumptions. While F1(H) drops marginally to 60.5, macro-F1 improves significantly to 65.7. This suggests that causality checking effectively suppresses noisy or irrelevant assumptions, resulting in a more balanced and robust system.



\section{Conclusion}

This work introduces the task of half-truth detection, addressing claims that are factually correct but misleading due to omitted context. To support this, we introduce \textsc{PolitiFact-Hidden}, a new benchmark with annotated evidence alignment and intent. We propose \textbf{TRACER}, a novel framework that detects omission-based misinformation via intent modeling and causal reasoning over hidden content. Integrated with existing fact verification models, TRACER consistently improves performance, especially on half-truths, demonstrating the importance of reasoning about omitted information. This work highlights omission-aware verification as a critical next step for building trustworthy fact-checking systems, and establishes TRACER as a generalizable framework for tackling this underexplored but essential challenge.

\newpage

\section*{Limitations}

While TRACER demonstrates strong performance in identifying omission-based misinformation, several limitations remain. First, our evaluation focuses on political discourse, as \textsc{PolitiFact-Hidden} is constructed from the PolitiFact corpus. While TRACER is designed to be model-agnostic and domain-independent, its effectiveness in other domains, such as health or finance, remains to be validated, especially where omission patterns may differ. Second, TRACER assumes that each claim expresses a coherent and inferable intent. However, real-world claims may be vague, ambiguous, or convey multiple overlapping intents, which can introduce noise in downstream reasoning. Future work may explore more robust modeling of claim pragmatics and intent uncertainty to extend TRACER's applicability to broader scenarios.

\section*{Acknowledgments}
This research is supported by the Ministry of Education, Singapore, under its MOE AcRF TIER 3 Grant (MOE-MOET32022-0001).

\bibliography{custom}

\begin{thebibliography}{27}
\providecommand{\natexlab}[1]{#1}

\bibitem[{Chen and Shu(2024)}]{survey_combating_misinformation}
Canyu Chen and Kai Shu. 2024.
\newblock \href {https://doi-org.libproxy1.nus.edu.sg/10.1002/aaai.12188}
  {Combating misinformation in the age of llms: Opportunities and challenges}.
\newblock \emph{AI Mag.}, 45(3):354–368.

\bibitem[{Chen et~al.(2022)Chen, Sriram, Choi, and Durrett}]{implicit-question}
Jifan Chen, Aniruddh Sriram, Eunsol Choi, and Greg Durrett. 2022.
\newblock \href {https://doi.org/10.18653/v1/2022.emnlp-main.229} {Generating
  literal and implied subquestions to fact-check complex claims}.
\newblock In \emph{Proceedings of the 2022 Conference on Empirical Methods in
  Natural Language Processing}, pages 3495--3516. Association for Computational
  Linguistics.

\bibitem[{Chong et~al.(2025)Chong, Tang, and Tung}]{DBLP:conf/emnlp/MPCG}
Brian Jun~Rong Chong, Yixuan Tang, and Anthony Kum~Hoe Tung. 2025.
\newblock Mpcg: Multi-round persona-conditioned generation for modeling the
  evolution of misinformation with llms.
\newblock In \emph{{EMNLP}}. Association for Computational Linguistics.

\bibitem[{Chung et~al.(2022)Chung, Hou, Longpre, Zoph et~al.}]{flan-t5}
Hyung~Won Chung, Le~Hou, Shayne Longpre, Barret Zoph, et~al. 2022.
\newblock \href {https://arxiv.org/abs/2210.11416} {Scaling
  instruction-finetuned language models}.
\newblock \emph{Preprint}, arXiv:2210.11416.

\bibitem[{Estornell et~al.(2020)Estornell, Das, and
  Vorobeychik}]{half_truth_deception}
Andrew Estornell, Sanmay Das, and Yevgeniy Vorobeychik. 2020.
\newblock \href {https://doi.org/10.1609/aaai.v34i06.6570} {Deception through
  half-truths}.
\newblock \emph{Proceedings of the AAAI Conference on Artificial Intelligence},
  34(06):10110--10117.

\bibitem[{Guo et~al.(2022)Guo, Schlichtkrull, and Vlachos}]{pipeline-survey}
Zhijiang Guo, Michael Schlichtkrull, and Andreas Vlachos. 2022.
\newblock \href {https://doi.org/10.1162/tacl_a_00454} {A survey on automated
  fact-checking}.
\newblock \emph{Transactions of the Association for Computational Linguistics},
  10:178--206.

\bibitem[{Holan(2018)}]{politifact_principles}
Angie~Drobnic Holan. 2018.
\newblock The principles of the truth-o-meter: Politifact’s methodology for
  independent fact-checking.
\newblock Last updated Jan. 12, 2024.

\bibitem[{Jaradat et~al.(2024)Jaradat, Zhang, and Li}]{cherry}
Israa Jaradat, Haiqi Zhang, and Chengkai Li. 2024.
\newblock \href {https://doi.org/10.48550/arXiv.2401.05650} {On detecting
  cherry-picking in news coverage using large language models}.
\newblock \emph{CoRR}, abs/2401.05650.

\bibitem[{Kojima et~al.(2022)Kojima, Gu, Reid, Matsuo, and
  Iwasawa}]{zeroshot-cot}
Takeshi Kojima, Shixiang~Shane Gu, Machel Reid, Yutaka Matsuo, and Yusuke
  Iwasawa. 2022.
\newblock Large language models are zero-shot reasoners.
\newblock In \emph{Proceedings of the 36th International Conference on Neural
  Information Processing Systems}, NIPS '22, Red Hook, NY, USA. Curran
  Associates Inc.

\bibitem[{Lee and Lee(2024)}]{overstatement_concealment}
Jiyoung Lee and Keeheon Lee. 2024.
\newblock \href {https://arxiv.org/abs/2408.00156} {Measuring falseness in news
  articles based on concealment and overstatement}.
\newblock \emph{Preprint}, arXiv:2408.00156.

\bibitem[{Liu et~al.(2019)Liu, Ott, Goyal, Du, Joshi, Chen, Levy, Lewis,
  Zettlemoyer, and Stoyanov}]{roberta-large}
Yinhan Liu, Myle Ott, Naman Goyal, Jingfei Du, Mandar Joshi, Danqi Chen, Omer
  Levy, Mike Lewis, Luke Zettlemoyer, and Veselin Stoyanov. 2019.
\newblock \href {https://arxiv.org/abs/1907.11692} {Roberta: {A} robustly
  optimized {BERT} pretraining approach}.
\newblock \emph{CoRR}, abs/1907.11692.

\bibitem[{Molina et~al.(2019)Molina, Sundar, Le, and Lee}]{completeness}
Maria~D. Molina, S.~Shyam Sundar, Thai Le, and Dongwon Lee. 2019.
\newblock \href {https://doi.org/10.1177/0002764219878224} {"fake news" is not
  simply false information: A concept explication and taxonomy of online
  content}.
\newblock \emph{American Behavioral Scientist}, 65(2):180--212.
\newblock Original work published 2021.

\bibitem[{Nils~Reimers(2019)}]{similarity_ranking_model}
Iryna~Gurevych Nils~Reimers. 2019.
\newblock Sentence-bert: Sentence embeddings using siamese bert-networks.
\newblock \emph{https://arxiv.org/abs/1908.10084}.

\bibitem[{Pan et~al.(2023{\natexlab{a}})Pan, Lu, Kan, and Nakov}]{qacheck}
Liangming Pan, Xinyuan Lu, Min-Yen Kan, and Preslav Nakov. 2023{\natexlab{a}}.
\newblock \href {https://doi.org/10.18653/v1/2023.emnlp-demo.23} {{QAC}heck: A
  demonstration system for question-guided multi-hop fact-checking}.
\newblock In \emph{Proceedings of the 2023 Conference on Empirical Methods in
  Natural Language Processing: System Demonstrations}, pages 264--273,
  Singapore. Association for Computational Linguistics.

\bibitem[{Pan et~al.(2023{\natexlab{b}})Pan, Wu, Lu, Luu, Wang, Kan, and
  Nakov}]{pan-etal-2023-fact}
Liangming Pan, Xiaobao Wu, Xinyuan Lu, Anh~Tuan Luu, William~Yang Wang, Min-Yen
  Kan, and Preslav Nakov. 2023{\natexlab{b}}.
\newblock \href {https://aclanthology.org/2023.acl-long.386/} {Fact-checking
  complex claims with program-guided reasoning}.
\newblock In \emph{Proceedings of the 61st Annual Meeting of the Association
  for Computational Linguistics (Volume 1: Long Papers)}, pages 6981--7004.

\bibitem[{Rodríguez-Ferrándiz(2023)}]{definition}
Raúl Rodríguez-Ferrándiz. 2023.
\newblock \href {https://doi.org/10.17645/mac.v11i2.6315} {An overview of the
  fake news phenomenon: From untruth-driven to post-truth-driven approaches}.
\newblock \emph{Media and Communication}, 11(2):15--29.

\bibitem[{Schlichtkrull et~al.(2023)Schlichtkrull, Guo, and
  Vlachos}]{schlichtkrull2023averitec}
Michael~Sejr Schlichtkrull, Zhijiang Guo, and Andreas Vlachos. 2023.
\newblock \href {https://openreview.net/forum?id=fKzSz0oyaI} {{AV}eri{T}e{C}: A
  dataset for real-world claim verification with evidence from the web}.
\newblock In \emph{Thirty-seventh Conference on Neural Information Processing
  Systems Datasets and Benchmarks Track}.

\bibitem[{Si et~al.(2024)Si, Zhao, Zhu, Zhu, Lu, and
  Zhou}]{si-etal-2024-checkwhy}
Jiasheng Si, Yibo Zhao, Yingjie Zhu, Haiyang Zhu, Wenpeng Lu, and Deyu Zhou.
  2024.
\newblock \href {https://doi.org/10.18653/v1/2024.acl-long.835} {{CHECKWHY}:
  Causal fact verification via argument structure}.
\newblock In \emph{Proceedings of the 62nd Annual Meeting of the Association
  for Computational Linguistics (Volume 1: Long Papers)}, pages 15636--15659.
  Association for Computational Linguistics.

\bibitem[{Singamsetty et~al.(2023)Singamsetty, Madaan, Mehta, Bhatnagar, and
  Bhattacharyya}]{beware_half_truth}
Sandeep Singamsetty, Nishtha Madaan, Sameep Mehta, Varad Bhatnagar, and Pushpak
  Bhattacharyya. 2023.
\newblock \href {https://arxiv.org/abs/2308.07973} {"beware of deception":
  Detecting half-truth and debunking it through controlled claim editing}.
\newblock \emph{Preprint}, arXiv:2308.07973.

\bibitem[{Tang et~al.(2021)Tang, Ng, and Tung}]{DBLP:conf/eacl/TangNT21}
Yixuan Tang, Hwee~Tou Ng, and Anthony K.~H. Tung. 2021.
\newblock Do multi-hop question answering systems know how to answer the
  single-hop sub-questions?
\newblock In \emph{{EACL}}, pages 3244--3249. Association for Computational
  Linguistics.

\bibitem[{Tang et~al.(2025)Tang, Shi, Sun, and Tung}]{DBLP:conf/emnlp/NEWSCOPE}
Yixuan Tang, Yuanyuan Shi, Yiqun Sun, and Anthony Kum~Hoe Tung. 2025.
\newblock Uncovering the bigger picture: Comprehensive event understanding via
  diverse news retrieval.
\newblock In \emph{{EMNLP}}. Association for Computational Linguistics.

\bibitem[{Thorne et~al.(2018)Thorne, Vlachos, Christodoulopoulos, and
  Mittal}]{thorne-etal-2018-fever}
James Thorne, Andreas Vlachos, Christos Christodoulopoulos, and Arpit Mittal.
  2018.
\newblock {FEVER}: a large-scale dataset for fact extraction and verification.
\newblock In \emph{Proceedings of the 2018 Conference of the North American
  Chapter of the Association for Computational Linguistics: Human Language
  Technologies}, pages 809--819.

\bibitem[{Wang et~al.(2024)Wang, Li, Li, Fu, Pei, and Wang}]{intent-query}
Bing Wang, Ximing Li, Changchun Li, Bo~Fu, Songwen Pei, and Shengsheng Wang.
  2024.
\newblock \href {https://doi.org/10.1145/3627673.3679799} {Why misinformation
  is created? detecting them by integrating intent features}.
\newblock In \emph{Proceedings of the 33rd ACM International Conference on
  Information and Knowledge Management}, CIKM '24, page 2304–2314, New York,
  NY, USA. Association for Computing Machinery.

\bibitem[{Wang(2017)}]{wang-2017-liar}
William~Yang Wang. 2017.
\newblock "liar, liar pants on fire": A new benchmark dataset for fake news
  detection.
\newblock In \emph{Proceedings of the 55th Annual Meeting of the Association
  for Computational Linguistics (Volume 2: Short Papers)}, pages 422--426.

\bibitem[{Yang et~al.(2024)Yang, Zhang, Gao, and Zhang}]{intent-matrix}
Chang Yang, Peng Zhang, Hui Gao, and Jing Zhang. 2024.
\newblock \href {https://doi.org/10.18653/v1/2024.emnlp-main.256} {Deciphering
  rumors: A multi-task learning approach with intent-aware hierarchical
  contrastive learning}.
\newblock In \emph{Proceedings of the 2024 Conference on Empirical Methods in
  Natural Language Processing}, pages 4471--4483. Association for Computational
  Linguistics.

\bibitem[{Yang et~al.(2022)Yang, Ma, Chen, Lin, Luo, and
  Chang}]{rawfc-yang-etal-2022-coarse}
Zhiwei Yang, Jing Ma, Hechang Chen, Hongzhan Lin, Ziyang Luo, and Yi~Chang.
  2022.
\newblock \href {https://aclanthology.org/2022.coling-1.230/} {A coarse-to-fine
  cascaded evidence-distillation neural network for explainable fake news
  detection}.
\newblock In \emph{Proceedings of the 29th International Conference on
  Computational Linguistics}, pages 2608--2621. International Committee on
  Computational Linguistics.

\bibitem[{Zhang and Gao(2023)}]{hiss}
Xuan Zhang and Wei Gao. 2023.
\newblock \href {https://doi.org/10.18653/v1/2023.ijcnlp-main.64} {Towards
  {LLM}-based fact verification on news claims with a hierarchical step-by-step
  prompting method}.
\newblock In \emph{Proceedings of the 13th International Joint Conference on
  Natural Language Processing and the 3rd Conference of the Asia-Pacific
  Chapter of the Association for Computational Linguistics (Volume 1: Long
  Papers)}, pages 996--1011, Nusa Dua, Bali. Association for Computational
  Linguistics.

\end{thebibliography}

\appendix

\newpage
\onecolumn

\section{Prompts for Constructing \textsc{Politifact-Hidden}}
\label{appendix:data}
This section presents the prompt templates employed in building the \textsc{Politifact-Hidden} dataset.

\begin{tcolorbox}[colback=gray!5, colframe=gray!50!black, title=Prompt: Evidence Relevance Classification]

You are tasked to determine the relevance of an evidence to an event.\\
You will be given a claim, the fact-checking justification of this claim, and an evidence. Is the evidence irrelevant to the event?\\

Irrelevant: The evidence does not talk about one aspect of the event.\\
Relevant: The evidence talks about one aspect of the event even if it does not directly address the claim or shares the general topics of the event or simply reference to the original claim.\\

You do not need to focus on does the evidence support or refute the claim.

\textbf{Evidence:} \texttt{\{evidence\}}\\
\textbf{Justification:} \texttt{\{ruling\}}\\
\textbf{Claim:} \texttt{\{claim\}}\\

Is the evidence relevant to the event?\\
A. Yes\\
B. No\\
Answer only one letter:

\end{tcolorbox}

\begin{tcolorbox}[colback=gray!5, colframe=gray!50!black, title=Prompt: Evidence Presence Classification]

You are tasked to determine whether the evidence is presented in a claim.\\
You will be given a claim and evidence. Is the evidence presented in the claim?\\

Presented should satisfy the following conditions: 
\begin{enumerate}
  \item The evidence partly or fully supports the claim. No contradiction is found.
  \item The evidence supports the claim without further reasoning, because information is \textbf{directly} and \textbf{explicitly} stated in the claim.
\end{enumerate}

\textbf{Evidence:} \texttt{\{evidence\}}\\
\textbf{Claim:} \texttt{\{claim\}}\\

Is the evidence presented in the claim?\\
A. Yes\\
B. No\\
Answer only one letter:

\end{tcolorbox}

\newpage

\begin{tcolorbox}[colback=gray!5, colframe=gray!50!black, title=Prompt: Enrich Fact-Checking Ruling with Evidence Given.]

You will be provided with the ruling and evidence from a fact-checking article. Your task is to enhance the clarity and depth of the ruling.\\

\textbf{Definitions:}
\begin{itemize}
  \item \textbf{Ruling:} A concise summary of the fact-checking article that includes the veracity rating of the claim.
  \item \textbf{Evidence:} The supporting details and collected data related to the claim.
\end{itemize}

\textbf{Requirements:}
\begin{itemize}
  \item \textbf{Identify Ambiguities:} Review the ruling and evidence to pinpoint any unclear or incomplete information in the ruling.
  \item \textbf{Enrich with Evidence:} Refer to the relevant parts of the evidence to expand the ruling. Ensure the enriched ruling explicitly explains how the evidence supports or contradicts the claim and connects directly to its veracity rating.
  \item \textbf{Create a Comprehensive Ruling:} The enhanced ruling should independently present the full context of the fact-checking process and the rationale for the given rating.
\end{itemize}

\textbf{Evidence:} \texttt{\{evidence\}}\\
\textbf{Ruling:} \texttt{\{ruling\}}\\

Do not output other thing except your enhanced ruling.

\end{tcolorbox}

\begin{tcolorbox}[colback=gray!5, colframe=gray!50!black, title=Prompt: Intent Analysis]

A claim would convey implicit intents. You are required to determine the intent of a claim based on context in Ruling.

\textbf{Definition:}
\begin{itemize}
  \item \textbf{Claim:} The claim that is checked.
  \item \textbf{Ruling:} Text to determine veracity and explain how the claim would shape people's understanding.
  \item \textbf{Intent:} The understanding of the event that the speaker wants to shape, which is not directly presented in the claim.
\end{itemize}

(3 Examples are omitted)

\textbf{Requirements:}
\begin{enumerate}
  \item Intent must be checkable. For example, "people should do something" is not checkable because it does not happen until now.
  \item Output intent in \texttt{<>}.
  \item Please think step by step. First write your rationale, then the intent.
\end{enumerate}

\textbf{Claim:} \texttt{\{claim\}}\\
\textbf{Ruling:} \texttt{\{ruling\}}\\

\end{tcolorbox}

To avoid repetition, we use colors in the prompts to denote different evaluation dimensions, which are assessed independently in practice.

\begin{tcolorbox}[colback=gray!5, colframe=gray!50!black, title=Generated Intent Evaluation (4 dimensions)]

You are required to determine whether the intended conclusion is \textcolor{red}{a plausible intent of the claim} / \textcolor{blue}{conveys the implicit meaning of the claim} /  \textcolor{orange}{readable} / \textcolor{cyan}{sufficient, meaning that it is understandable within the scope of general knowledge}.

\textbf{Please rate using the following scale:}
\begin{itemize}
  \item \textcolor{red}{\textbf{0 (not plausible):} The claim contradicts the intended conclusion.}
  \item \textcolor{red}{\textbf{1 (plausible):} The claim does not contradict the intended conclusion.}
   \item  \textcolor{blue}{\textbf{0 (not implicit):} The intended conclusion simply rephrases some part of the claim. It does not convey any implicit meaning of the claim.}
  \item \textcolor{blue}{\textbf{1 (implicit):} The intended conclusion reveals implicit information that is not explicitly stated in the claim.}
     \item \textcolor{orange}{\textbf{0 (not readable):} The intended conclusion is not readable and is overly complicated.}
   \item \textcolor{orange}{\textbf{1 (readable):} The intended conclusion is readable and understandable.}
  \item \textcolor{cyan}{\textbf{0 (not sufficient):} The intended conclusion has obvious ambiguous references and is not understandable. For example, it uses unclear terms like “the claim”.}
  \item \textcolor{cyan}{\textbf{1 (sufficient):} The intended conclusion is clearly referenced and understandable on its own.}

\end{itemize}

\textbf{Claim:} \texttt{\{claim\}}\\
\textbf{Intended Conclusion:} \texttt{\{intent\}}\\

Output only one digit.

\end{tcolorbox}

\clearpage

\section{Prompt Templates Used in TRACER}
\label{sec:appendix_prompts}

This appendix provides the complete prompt templates employed at each stage of the \textsc{TRACER} framework. We include prompts for implicit question generation, assumption inference, causality evaluation, and final re-assessment.

\begin{tcolorbox}[colback=gray!5, colframe=gray!50!black, title=Prompt: Implicit Questions Generation.]

A claim can be literally accurate but still misleading in an implicit way.\\
Your task is to identify the important implicit questions addressed by the evidence.\\

\textbf{Steps:}
\begin{enumerate}
  \item Read the evidence below carefully to understand the full context and the topics it covers.
  \item Assume the claim is true. What important implicit yes-no questions should be asked to verify the intended conclusion, rather than just the literal accuracy of the claim?
  \item Generate 1--3 such implicit questions.
  \item Each question should be enclosed in its own angle brackets \texttt{<>}.
  \item All implicit questions must be yes-no questions.
\end{enumerate}

(Examples are omitted.)

\textbf{Claim:} \texttt{\{claim\}} \\
\textbf{Intended conclusion:} \texttt{\{intent\}} \\
\textbf{Evidence:} \\
\texttt{\{evidence\}}

\end{tcolorbox}

\newpage

\begin{tcolorbox}[colback=gray!5, colframe=gray!50!black, title=Prompt: Assumption Generation]

A claim could be literally accurate but still misleading because of its intended conclusion.\\
Your task is to determine what assumptions the intended conclusion is based on, besides the claim.

\textbf{Definition:}
\begin{itemize}
  \item \textbf{Claim:} A statement assumed to be true.
  \item \textbf{Intended conclusion:} The intended conclusion of the claim, which needs checking.
  \item \textbf{Questions:} Some important questions when checking the claim.
  \item \textbf{Assumptions:} The assumptions that the intended conclusion is based on, besides the claim.
\end{itemize}

\textbf{Steps:}
\begin{enumerate}
  \item Read the claim, intended conclusion, and questions.
  \item Assuming the claim is correct, what assumptions does the question imply should serve as the basis for the intended conclusion?
  \item Output a 1--3 sentence rationale, followed by 1--\texttt{\{assumption\_max\_number\}} assumptions.\\
        Each assumption should be enclosed in angle brackets \texttt{<>} and separated by \texttt{||}.
\end{enumerate}

\textbf{Requirements:}
\begin{enumerate}
  \item Ensure that each assumption can independently convey its meaning.\\
        For example, never use vague references like ``the claim,'' ``the evidence,'' or ``the intent''; instead, refer to specific information.
  \item Only include assumptions that you are confident in and that serve as a strong basis for the intended conclusion.
\end{enumerate}

(Examples are omitted.)

\textbf{Claim:} \texttt{\{claim\}} \\
\textbf{Intended conclusion:} \texttt{\{intention\}} \\
\textbf{Questions:} \\
\texttt{\{questions\}}

\end{tcolorbox}

\begin{tcolorbox}[colback=gray!5, colframe=gray!50!black, title=Prompt: Causality Analysis]

You are required to do a counterfactual causal inference on a given causal graph.

\textbf{Argument:}
\begin{verbatim}
{
  "Z": intent,
  "linked_by": {
    "X": claim,
    "Y_1": assumption_1,
    "Y_2": assumption_2
  }
}
\end{verbatim}

Evaluate $\Delta P(Z \mid \text{do}(\{\text{letter}\} = \neg\{\text{letter}\}))$.\\
More specifically, how does the probability of $Z$ change when we set \{\text{letter}\} from \{\text{letter}\} to $\neg$\{\text{letter}\}?

\textbf{Options:}

A. The probability of $Z$ does not change.

B. The probability of $Z$ increases (Z becomes more likely to be true).

C. The probability of $Z$ decreases (Z becomes less likely to be true).

Please answer with \textbf{one letter only.}

\end{tcolorbox}

\begin{tcolorbox}[colback=gray!5, colframe=gray!50!black, title=Prompt: Re-Assessment]

A claim may be factually accurate but still misleading due to its implied conclusion. Your task is to refine the veracity assessment of such a claim by considering additional hidden information.

You are given a previously generated fact-checking \textbf{justification}, along with new \textbf{evidence} and an \textbf{argument} supporting the intended conclusion of the claim.

Please determine whether the justification has already addressed the hidden information. Then, refine the veracity of the claim accordingly.

\textbf{Input:}
\begin{verbatim}
Evidence: [EVIDENCE]
Argument: [ARGUMENT]
Justification: [JUSTIFICATION]
\end{verbatim}

\textbf{Instruction:}  
Reassess the veracity of the claim based on the above.  

\textbf{Choose one of the following options (output only the letter):} 

A. True //
B. Half-true //
C. False //
D. Unverifiable (e.g., the hidden assumption does not support the conclusion, or the information is insufficient)

\textbf{Your answer (one letter only):}

\end{tcolorbox}

\section{Generalization with LLaMA2-7B}

\label{app:llama2}
Results in Table~\ref{tab:llama2} demonstrate that TRACER yields consistent improvements across metrics when applied to the open-source \texttt{LLaMA2-7B}, with particularly notable gains in Half-True classification. 

\begin{table}[h!]
\centering
\begin{tabular}{lccccc}
\toprule
Model & Accuracy & Macro-F1 & F1(True) & F1(Half-True) & F1(False) \\
\midrule
HiSS  & 78.3 & 59.4 & 44.7 & 44.4 & 88.9 \\
HiSS + RA (GPT-3.5-turbo) & 81.9 & \textbf{65.7} & \textbf{46.6} & 60.5 & 90.1 \\
HiSS + RA (\texttt{LLaMA2-7B}) & \textbf{82.3} & 65.4 & 43.6 & \textbf{61.3} & \textbf{91.2} \\
\bottomrule
\end{tabular}
\caption{Generalization of TRACER with different backbones.}
\label{tab:llama2}
\end{table}

We further analyze TRACER’s performance across different claim lengths. Using the open-source \texttt{LLaMA2-7B} backbone, we partition test claims into four length ranges by word count. 
Table~\ref{tab:length} shows consistent improvements across all ranges, with larger gains observed for longer claims, 
which likely offer richer context for intent inference and assumption generation.

\begin{table}[h!]
\centering
\begin{tabular}{lcccc}
\toprule
\textbf{Model} & 4--13 (755) & 14--23 (893) & 24--34 (308) & $\geq$35 (44) \\
\midrule
HiSS         & 80.3 & 78.5 & 73.1 & 75.0 \\
HiSS + RA    & \textbf{83.7} & \textbf{81.8} & \textbf{80.5} & \textbf{81.8} \\
Improvement  & $\uparrow$3.4 & $\uparrow$3.3 & $\uparrow$7.5 & $\uparrow$6.8 \\
\bottomrule
\end{tabular}
\caption{TRACER’s performance across different claim lengths (F1 scores) using \texttt{LLaMA2-7B}. 
Numbers in parentheses indicate the number of examples per length range. Longer claims provide richer context, leading to larger improvements.}
\label{tab:length}
\end{table}

\end{document}